\documentclass{article}
\usepackage{natbib}
\setcitestyle{numbers}

\usepackage[dblblindworkshop,final]{neurips_2026}

\usepackage{hyperref}
\usepackage{url}

\usepackage{microtype}
\usepackage{graphicx}
\usepackage{adjustbox}
\usepackage{subcaption}
\usepackage{booktabs} 

\usepackage{amsmath}	
\usepackage{amssymb}	
\usepackage{booktabs}
\usepackage{times}
\usepackage{caption}
\usepackage{microtype}
\usepackage{epsfig}
\usepackage[table,xcdraw,dvipsnames]{xcolor}
\usepackage{caption}
\usepackage{float}
\usepackage{placeins}
\usepackage{color, colortbl}
\usepackage{stfloats}
\usepackage{enumitem}
\usepackage{tabularx}
\usepackage{xstring}
\usepackage{multirow}
\usepackage{xspace}
\usepackage{url}
\usepackage{subcaption}
\usepackage{booktabs}
\usepackage{multirow}
\usepackage{array}          
\usepackage[table]{xcolor}  
\usepackage{xcolor}
\usepackage{makecell}
\usepackage[hang,flushmargin]{footmisc}

\definecolor{trajshade}{RGB}{226,239,245}
\newcolumntype{Z}{>{\columncolor{trajshade}}c}

\newif\ifreview 
\newif\ifarxiv 
\newif\ifcamera 
\newif\ifrebuttal 

\ifcamera \usepackage[accsupp]{axessibility} \fi

\usepackage{amsthm}

\usepackage{subcaption}   
\usepackage{graphicx}
\usepackage{caption}
\ifarxiv  \fi

\usepackage{amsmath}
\usepackage{amssymb}
\usepackage{mathtools}
\usepackage{amsthm}
\usepackage{mdframed}
\usepackage{tikz}
\usepackage{etoolbox}
\usepackage[capitalize,noabbrev]{cleveref}
\usepackage{thmtools}

\theoremstyle{plain}
\newtheorem*{theorem*}{Theorem}

\usepackage{booktabs}

\workshoptitle{The 6th Workshop on Mathematical Reasoning and AI}

\usepackage[utf8]{inputenc} 
\usepackage[T1]{fontenc}    
\usepackage{hyperref}       
\usepackage{url}            
\usepackage{booktabs}       
\usepackage{amsfonts}       
\usepackage{nicefrac}       
\usepackage{microtype}      
\usepackage{xcolor}         

\title{Preserving Mathematical Reasoning in Compressed Diffusion Language Models via Trajectory-Aware Low-Rank Approximation}

\author{%
  Tian Liang$^1$, \quad Zishan Shao$^{1,\thanks{Correspondence to: zishan.shao@duke.edu.}}$, \quad Yiran Chen$^1$ \\
  $^1$Department of Electrical and Computer Engineering, Duke University \\
  \texttt{\{tian.liang, zishan.shao, yiran.chen\}@duke.edu} \\
}

\begin{document}

\maketitle

\begin{abstract}

    Diffusion language model (dLLM) compression faces a known challenge because calibration is typically performed on clean, fully visible activations, whereas inference traverses partially masked intermediate states. For low-rank compression, this raises two questions. First, can low-rank optimality still be characterized when approximation quality is measured over trajectory-distributed states, and second, does the choice of calibration states affect mathematical reasoning preservation under compression? We address these questions by formulating a trajectory-aware low-rank objective over corruption levels and masking realizations. To estimate this objective efficiently, we propose Traj-MC, which estimates the trajectory second moment through Monte Carlo sampling and yields exact sampled-state optimality and population consistency. Under matched compression budgets, trajectory-aware calibration improves reconstruction over the generation trajectory and preserves substantially more mathematical reasoning than clean calibration on mathematical reasoning benchmarks. Our results connect trajectory-aware low-rank optimality to the reasoning capability retained after dLLM compression. Our code is available at: \url{https://github.com/Zishan-Shao/traj-mc.git}.

\end{abstract}

\section{Introduction}

Low-rank compression~\cite{lin2025modegpt,yuan2023asvd,wang2025svdllm} offers a promising path toward more efficient diffusion language models (dLLMs)~\cite{nie2025llada,wang2026dream}, but dLLM compression faces a known calibration challenge. Activation-aware low-rank methods~\cite{yuan2023asvd,wang2025svdllm} typically estimate which directions to preserve from clean, fully visible sequences, whereas inference traverses partially observed states during iterative denoising~\cite{zhang2026quantdllm,li2023qdiffusion,xu2025dllmquant}. For low-rank compression, this raises two questions: whether low-rank optimality can still be characterized over trajectory states, and whether the choice of calibration states affects mathematical reasoning preservation.

\begin{figure}[t]
    \centering
    \includegraphics[width=0.99\linewidth]{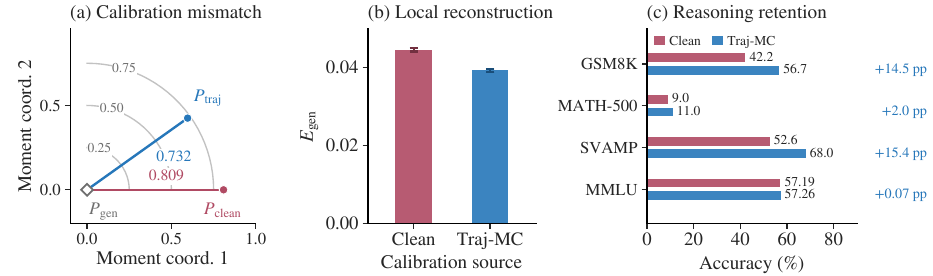}
    \caption{
    \textbf{Impact of trajectory-aware calibration.} 
    (a) Traj-MC reduces second-moment mismatch to generation states (details in Appendix.~\ref{app:exp}), 
    (b) lowers held-out reconstruction error $E_{\text{gen}}$, and 
    (c) preserves substantially higher mathematical reasoning performance at 20\% compression with similar understanding accuracy.
    }
    \label{fig:overview}
\end{figure}

We study these questions from a distributional perspective. Activation-aware low-rank approximation is defined relative to the states over which reconstruction error is measured, so changing the calibration distribution can change the optimal low-rank solution~\cite{wang2025svdllm}. We therefore formulate a trajectory-aware objective over corruption levels and masking realizations, shifting the question from how SVD is computed to \textbf{\textit{which state distribution should define the geometry of the approximation.}} Traj-MC estimates this objective from corrupted calibration windows through Monte Carlo sampling. As shown in Fig.~\ref{fig:overview}(a), trajectory-aware calibration better matches the second moment of generation states than clean calibration.

This improved alignment carries over to both local fidelity and downstream behavior. Figure~\ref{fig:overview}(b) shows lower reconstruction error on held-out generation states, while
Fig.~\ref{fig:overview}(c) shows gains on the displayed mathematical
reasoning benchmarks at 20\% parameter reduction, with MMLU remaining nearly unchanged. Together, these results connect the state distribution defining low-rank optimality to the reasoning capability retained after compression.

To our knowledge, this is the first study of low-rank compression for dLLMs. Our contributions are threefold: (1) we formulate a trajectory-aware low-rank objective and establish the corresponding fixed-rank optimal solution over trajectory-distributed states; (2) we develop Traj-MC, a Monte Carlo approximation to this population objective with consistency guarantees; and (3) we quantify the behavioral consequences of trajectory-aware approximation, showing improved mathematical reasoning preservation under matched compression budgets.


\section{Related Work}

\paragraph{Activation-aware low-rank compression.}
Activation-aware methods improve low-rank approximation by incorporating calibration activations. ASVD uses activation statistics, SVD-LLM aligns truncation with activation reconstruction through whitening, and MoDeGPT extends reconstruction across coupled Transformer modules \citep{yuan2023asvd,wang2025svdllm,lin2025modegpt}. These works show that low-rank quality depends on the activations defining reconstruction. We ask a complementary question for dLLMs~\cite{nie2025llada,wang2026dream}: \textbf{\textit{which activation distribution should define the low-rank objective when inference traverses multiple intermediate states?}}

\paragraph{Compression under iterative generation.}
Timestep-dependent calibration mismatch is well recognized in diffusion-model compression. Diffusion quantization methods use timestep-aware calibration \citep{li2023qdiffusion,wang2024accurateptq}, while DLLMQuant and Quant-dLLM explicitly model masked denoising states for dLLM quantization \citep{xu2025dllmquant,zhang2026quantdllm}. We instead study what this known mismatch implies for \textbf{\textit{low-rank optimality}}, deriving a trajectory-aware objective and fixed-rank solution, approximating it with Traj-MC, and evaluating its consequences for mathematical reasoning preservation.

\section{Methodology}
\label{sec:method}


We instantiate Traj-MC with the whitening-based SVD solver of
SVD-LLM~\cite{wang2025svdllm}; the resulting method is denoted
+Traj-MC in the tables.

\subsection{Trajectory-aware low-rank optimality}
\label{sec:trajectory_optimality}

Consider a linear layer
$W\in\mathbb R^{d_{\mathrm{out}}\times d_{\mathrm{in}}}$
in a fixed reference model. For a length-$L$ calibration window
$x\sim\mathcal D$, corruption level $t\sim\pi$, and mask
$m\sim q(\cdot\mid t)$, let
$h(x,t,m)\in\mathbb R^{d_{\mathrm{in}}\times L}$
denote the input activations.
These corrupted windows provide a tractable surrogate for
partially observed denoising states. We seek a single rank-$k$
operator under the integrated reconstruction objective
\begin{equation}
\begin{aligned}
R_{\mathrm{traj}}(\widetilde W)
&=\mathbb E_{x,t,m}
\bigl[\|(W-\widetilde W)h(x,t,m)\|_F^2\bigr],\\
\Sigma_{\mathrm{traj}}
&=\mathbb E_{x,t,m}
\bigl[h(x,t,m)h(x,t,m)^\top\bigr].
\end{aligned}
\label{eq:trajectory_objective}
\end{equation}
The uncentered second moment $\Sigma_{\mathrm{traj}}$ defines
reconstruction geometry across examples, corruption levels,
and masking realizations.
For $N$ sampled windows $h_i=h(x_i,t_i,m_i)$, let
$H_N=[h_1,\ldots,h_N]$ and define
\begin{equation}
\widehat R_N(\widetilde W)
=\frac{1}{N}\|(W-\widetilde W)H_N\|_F^2,
\qquad
\widehat\Sigma_N=\frac{1}{N}H_NH_N^\top.
\label{eq:empirical_trajectory}
\end{equation}

\paragraph{Proposition 1 (Exact sampled-state optimality).}
\label{eq:prop1}
If $\widehat\Sigma_N\succ0$, with Cholesky factorization
$\widehat\Sigma_N=S_NS_N^\top$, then
\begin{equation}
\widehat W_{N,k}^{\star}
=(WS_N)_kS_N^{-1}
\in\underset{\operatorname{rank}(\widetilde W)\le k}
{\arg\min}\;\widehat R_N(\widetilde W),
\label{eq:empirical_optimality}
\end{equation}
where $(\cdot)_k$ denotes a best rank-$k$ SVD truncation.
The identity
$\widehat R_N(\widetilde W)=\|(W-\widetilde W)S_N\|_F^2$
reduces the problem to rank-$k$ approximation of $WS_N$.
This guarantee is conditional on the collected states and
requires no independence assumptions.

\subsection{Traj-MC: Monte Carlo approximation}
\label{sec:traj_mc}

Traj-MC samples $t\sim\mathcal U(0,1)$ and, conditional on $t$,
masks each token independently with probability $t$, avoiding
full dense-model rollouts. It accumulates $\widehat\Sigma_N$
and applies Eq.~\eqref{eq:empirical_optimality}, leaving the
compressed layers and rank allocation unchanged.
For analysis, we assume i.i.d. window tuples from
$\mathcal D(x)\pi(t)q(m\mid t)$; tokens within a window need
not be independent.
Under $\mathbb E\|h\|_F^4<\infty$, the estimator is unbiased and
$\mathbb E\|\widehat\Sigma_N-\Sigma_{\mathrm{traj}}\|_F^2
=O(N^{-1})$.

\paragraph{Proposition 2 (Population consistency).}
Assume $\mathbb E\|h\|_F^2<\infty$ and
$\Sigma_{\mathrm{traj}}=S_{\mathrm{traj}}S_{\mathrm{traj}}^\top
\succ0$, with $S_{\mathrm{traj}}$ its Cholesky factor.
If $\sigma_k(WS_{\mathrm{traj}})
>\sigma_{k+1}(WS_{\mathrm{traj}})$, then by the strong law of large numbers,
\begin{equation}
\begin{aligned}
\widehat\Sigma_N
&\xrightarrow{\mathrm{a.s.}}\Sigma_{\mathrm{traj}},\\
\widehat W_{N,k}^{\star}
&\xrightarrow{\mathrm{a.s.}}
W_{\mathrm{traj},k}^{\star}
=(WS_{\mathrm{traj}})_kS_{\mathrm{traj}}^{-1}.
\end{aligned}
\label{eq:population_consistency}
\end{equation}
Here $W_{\mathrm{traj},k}^{\star}$ globally minimizes
$R_{\mathrm{traj}}$ over operators of rank at most $k$.
Proofs and the variance derivation appear in
Appendix~\ref{app:trajectory_theory}.


\section{Experiments}
\label{sec:experiments}

\subsection{Experimental Setup}
\label{sec:setup}

We evaluate LLaDA-8B-Instruct~\cite{nie2025llada}, and Dream-7B-Instruct~\cite{wang2026dream}
at 20\% and 40\% parameter reduction. Within each model, SVD-LLM and
Traj-MC use matched compression settings and the same C4 calibration
windows, differing only in whether the activations are collected from
clean or trajectory-masked states. Vanilla Weight SVD serves as a no-activation
reference. Each model follows its official evaluation protocol. 

\subsection{Main Results}
\label{sec:main-results}

\paragraph{Trajectory-aware calibration better preserves reasoning.}
Table~\ref{tab:main_results} shows that changing only the calibration state distribution substantially alters post-compression behavior. Across mathematical and reasoning benchmarks (GSM8K, MATH-500, SVAMP, ARC-C), Traj-MC consistently outperforms clean-state calibration on both dLLMs, with the advantage persisting under more aggressive compression ratios. 
This positive trend generalizes across both instruct and base model variants, whereas performance on knowledge and commonsense benchmarks exhibits relatively minor or mixed changes. 
These findings indicate that calibration state alignment is particularly critical for preserving complex reasoning capabilities under low-rank compression.

\paragraph{Local trajectory fidelity transfers to actual generation states.}
The downstream gains coincide with improved reconstruction error ($E_{\mathrm{gen}}$) on states encountered during real denoising. 
Using only corrupted calibration windows, Traj-MC consistently lowers generation-state error compared to clean-state calibration. 
This empirically bridges the distributional objective in Sec.~\ref{sec:trajectory_optimality} to the actual deployment states of the compressed model.

\begin{figure}[t]
    \centering
    \includegraphics[width=0.99\linewidth]{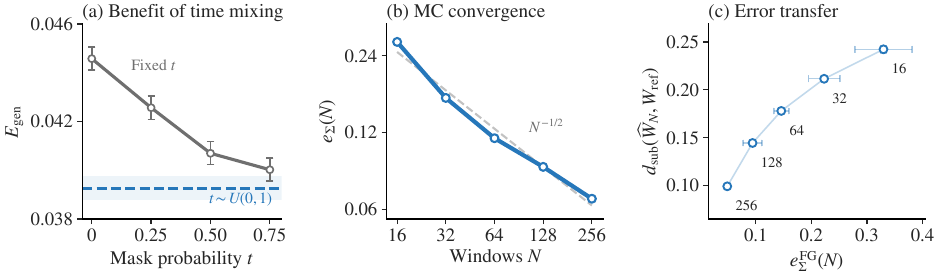}
    \caption{
    \textbf{Empirical validation of Traj-MC.} 
    (a) Uniform sampling achieves lower generation-state error $E_{\text{gen}}$ than any fixed-$t$ baseline. 
    (b) Estimation error $e_{\Sigma}(N)$ follows the theoretical $O(N^{-1/2})$ convergence rate. 
    (c) Covariance error directly tracks the resulting low-rank subspace deviation $d_{\text{sub}}$.
    }
    \label{fig:mc_analysis}
\end{figure}

\paragraph{Why trajectory mixing?}
Figure~\ref{fig:mc_analysis} isolates the two ingredients behind
Traj-MC. First, the advantage does not rely on selecting or tuning a specific corruption level. Increasing a fixed mask probability
progressively improves generation-state reconstruction, yet uniform
sampling across corruption levels remains better than every fixed-$t$
alternative (Fig.~\ref{fig:mc_analysis}a). This supports optimizing over
a distribution of intermediate states rather than calibrating to one
representative timestep.

Second, the Monte Carlo approximation behaves as predicted by the
analysis. The estimated second moment approaches its reference value
approximately at the expected $N^{-1/2}$ rate
(Fig.~\ref{fig:mc_analysis}b), and improvements in covariance estimation
are accompanied by smaller deviations of the resulting low-rank
solution (Fig.~\ref{fig:mc_analysis}c). Together, these results indicate
that Traj-MC's gains reflect both the trajectory-level target itself and
a stable Monte Carlo approximation to that target.

\definecolor{trajshade}{RGB}{226,239,245}
\newcommand{\trajcell}[1]{\cellcolor{trajshade}#1}

\begin{table*}[t]
\centering
\caption{Effect of calibration distribution on low-rank dLLM compression under matched settings, where \textbf{bold} marks Traj-MC improvements over SVD-LLM (higher is better).}
\label{tab:main_results}

\small
\setlength{\tabcolsep}{3.2pt}
\renewcommand{\arraystretch}{1.08}

\begin{tabular}{l l l cccc ccc}
\toprule
\multirow{2}{*}{Model}
& \multirow{2}{*}{Setting}
& \multirow{2}{*}{Method}
& \multicolumn{4}{c}{Math \& Reasoning}
& \multicolumn{3}{c}{General Understanding} \\
\cmidrule(lr){4-7}
\cmidrule(lr){8-10}
& & & GSM8K & MATH-500 & SVAMP & ARC-C & MMLU & HellaSwag & PIQA \\
\midrule


 & Dense & BF16 & 80.2 & 30.2 & 85.7 & 87.9 & 64.3 & 76.6 & 82.8 \\
\cmidrule(lr){2-10}
 & & SVD & 12.3 & 4.8 & 28.8 & 42.3 & 39.7 & 38.2 & 55.9 \\
 & & SVD-LLM & 42.2 & 9.0 & 52.6 & 77.8 & 57.2 & 65.6 & 78.7 \\
\rowcolor{trajshade}
\cellcolor{white} & \cellcolor{white}\multirow{-3}{*}{20\%} & +Traj-MC
 & \textbf{56.7} & \textbf{11.0} & \textbf{68.0} & \textbf{79.9} & \textbf{57.3} & 65.4 & 77.7 \\
\cmidrule(lr){2-10}
 & & SVD & 0.0 & 0.2 & 0.0 & 16.7 & 0.0 & 0.1 & 0.4 \\
 & & SVD-LLM & 4.2 & 2.0 & 13.9 & 54.2 & 45.8 & 55.4 & 72.5 \\
\rowcolor{trajshade}
\cellcolor{white}\multirow{-7}{*}{\begin{tabular}[c]{@{}l@{}}LLaDA-8B\\Instruct\end{tabular}}
 & \cellcolor{white}\multirow{-3}{*}{40\%} & +Traj-MC
 & \textbf{11.2} & \textbf{2.4} & \textbf{18.7} & \textbf{65.3} & \textbf{48.5} & 52.6 & 68.7 \\
\midrule


 & Dense & BF16 & 83.0 & 40.6 & 83.7 & 71.1 & 72.3 & 52.9 & 69.5 \\
\cmidrule(lr){2-10}
 & & SVD & 0.8 & 1.0 & 1.2 & 0.0 & 0.3 & 26.4 & 52.7 \\
 & & SVD-LLM & 7.6 & 5.0 & 8.0 & 28.1 & 42.4 & 41.6 & 61.0 \\
\rowcolor{trajshade}
\cellcolor{white}\multirow{-4}{*}{\begin{tabular}[c]{@{}l@{}}Dream-7B\\Instruct\end{tabular}}
 & \cellcolor{white}\multirow{-3}{*}{20\%} & +Traj-MC
 & \textbf{30.7} & \textbf{8.4} & \textbf{37.0} & \textbf{33.8} & 38.4 & \textbf{44.9} & \textbf{63.7} \\
\bottomrule
\end{tabular}
\end{table*}






\section{Conclusion}


We formulate low-rank dLLM compression over trajectory-distributed states and derive the corresponding exact fixed-rank optimum. Traj-MC turns this population objective into a practical Monte Carlo approximation without requiring dense generation rollouts. Under matched compression settings, trajectory-aware calibration improves reconstruction on held-out generation states and preserves substantially more mathematical reasoning than clean-state calibration, while changes on knowledge and commonsense tasks are comparatively smaller or mixed. These results show that low-rank compression quality in dLLMs depends not only on the decomposition rule, but also on the state distribution used to define its reconstruction geometry. More broadly, they identify calibration-state design as a first-class component of low-rank dLLM compression.

\definecolor{trajshade}{RGB}{226,239,245}
\newcolumntype{Z}{>{\columncolor{trajshade}}c}
\newcommand{\holm}{\rlap{$^{\ast}$}}    
\newcommand{\raw}{\rlap{$^{\dagger}$}}  

\bibliography{refs}
\bibliographystyle{unsrtnat}


\appendix


\section{Theoretical Details}
\label{app:trajectory_theory}

\subsection{Exact optimality for sampled trajectory states}
\label{app:sampled_optimality}

\paragraph{Proof of Proposition 1.}
Define the normalized, whitened activations
$Z_N=N^{-1/2}S_N^{-1}H_N$.
Since $N^{-1}H_NH_N^\top=S_NS_N^\top$, we have
$Z_NZ_N^\top=I$. Therefore, for every candidate $\widetilde W$,
\begin{equation}
\begin{aligned}
\widehat R_N(\widetilde W)
&=\|(W-\widetilde W)S_NZ_N\|_F^2\\
&=\operatorname{tr}\!\left[
(W-\widetilde W)S_NZ_NZ_N^\top
S_N^\top(W-\widetilde W)^\top\right]\\
&=\|(W-\widetilde W)S_N\|_F^2.
\end{aligned}
\label{eq:trajectory_whitening}
\end{equation}
Because $S_N$ is invertible, the change of variables
$A=\widetilde W S_N$ preserves rank. The Eckart--Young theorem
applied to $WS_N$ gives Eq.~\eqref{eq:empirical_optimality}
and the minimum reconstruction error
\begin{equation}
\widehat R_N(\widehat W_{N,k}^{\star})
=\sum_{j>k}\sigma_j^2(WS_N).
\label{eq:empirical_optimal_value}
\end{equation}
This argument is deterministic conditional on the collected
states; it does not require independence across windows or tokens.

\subsection{Monte Carlo variance decay}
\label{app:mc_variance}

Let $Y_i=h_ih_i^\top$ for i.i.d. window tuples and assume
\[
V_{\mathrm{traj}}
=\mathbb E\|Y_i-\Sigma_{\mathrm{traj}}\|_F^2<\infty.
\]
A sufficient condition is $\mathbb E\|h\|_F^4<\infty$.
Since $\widehat\Sigma_N=N^{-1}\sum_iY_i$, linearity of
expectation gives $\mathbb E[\widehat\Sigma_N]
=\Sigma_{\mathrm{traj}}$. Moreover,
\begin{equation}
\begin{aligned}
\mathbb E\|\widehat\Sigma_N-\Sigma_{\mathrm{traj}}\|_F^2
&=\frac{1}{N^2}\sum_{i,j=1}^{N}
\mathbb E\left\langle
Y_i-\Sigma_{\mathrm{traj}},
Y_j-\Sigma_{\mathrm{traj}}
\right\rangle_F\\
&=\frac{V_{\mathrm{traj}}}{N}.
\end{aligned}
\label{eq:mc_variance}
\end{equation}
Independence across windows and zero mean eliminate every
cross term with $i\ne j$. Thus the mean-squared Frobenius
estimation error decays as $N^{-1}$, and its root mean square
as $N^{-1/2}$. No independence among tokens within a window
is needed.

\subsection{Consistency of the trajectory-optimal solution}
\label{app:population_consistency}

\paragraph{Proof of Proposition 2.}
The bound $\|hh^\top\|_F\le\|h\|_F^2$ ensures that every
entry of $hh^\top$ is integrable. Applying the strong law
entrywise, and using finite dimensionality, gives
$\widehat\Sigma_N\xrightarrow{\mathrm{a.s.}}
\Sigma_{\mathrm{traj}}$ in matrix norm.
Positive definiteness of the limit implies that
$\widehat\Sigma_N$ is eventually positive definite almost surely.
Continuity of Cholesky factorization and inversion yields
$S_N\xrightarrow{\mathrm{a.s.}}S_{\mathrm{traj}}$ and
$S_N^{-1}\xrightarrow{\mathrm{a.s.}}S_{\mathrm{traj}}^{-1}$.
The strict spectral gap makes the best rank-$k$ truncation
continuous at $WS_{\mathrm{traj}}$, so
\[
(WS_N)_kS_N^{-1}
\xrightarrow{\mathrm{a.s.}}
(WS_{\mathrm{traj}})_kS_{\mathrm{traj}}^{-1}.
\]
Finally,
\[
R_{\mathrm{traj}}(\widetilde W)
=\operatorname{tr}\!\left[
(W-\widetilde W)\Sigma_{\mathrm{traj}}
(W-\widetilde W)^\top\right]
=\|(W-\widetilde W)S_{\mathrm{traj}}\|_F^2.
\]
The same rank-preserving change of variables as in
Proposition 1 establishes global population optimality.

\subsection{Why Uniform Corruption Mixing?}
\label{app:uniform_mixing}

We choose uniform corruption mixing to balance calibration across
corruption severities, measured by the number of masked tokens.
For a calibration window with $L$ maskable tokens, Traj-MC samples
$t\sim\mathrm{Unif}(0,1)$ once per window and, conditional on $t$,
masks each token independently with probability $t$.
Let $K$ denote the resulting number of masked tokens. Since
$K\mid t\sim\mathrm{Binomial}(L,t)$,
\[
\begin{aligned}
\Pr(K=j)
&=
\binom{L}{j}
\int_0^1 t^j(1-t)^{L-j}\,dt \\
&=
\frac{1}{L+1},
\qquad j=0,\ldots,L.
\end{aligned}
\]
Thus every mask count receives the same marginal probability.
Conditional on $K=j$, all subsets of $j$ token positions are equally
likely. For comparison, a fixed masking probability $t_0$ gives
$\operatorname{Var}(K/L)=t_0(1-t_0)/L$, so the mask fraction
concentrates around $t_0$ as the window length increases.

This property directly characterizes the weighting in our
reconstruction objective. By the law of total expectation,
\[
R_{\mathrm{traj}}(\widetilde W)
=
\frac{1}{L+1}
\sum_{j=0}^{L}
\mathbb E\!\left[
\|(W-\widetilde W)h(x,t,m)\|_F^2
\,\middle|\, K=j
\right].
\]
Uniform mixing therefore assigns equal weight to the
mask-count-conditioned reconstruction risks, rather than concentrating
the population objective around one masking severity.


\section{Experimental Details}
\label{app:exp}

\paragraph{Calibration protocol.}
All activation-aware methods use the same 256 C4 calibration windows,
compressed layers, and rank allocation; only the states used to collect
activations differ. Clean calibration uses fully visible windows, while
Traj-MC samples $t\sim\mathcal U(0,1)$ and independently masks each token
with probability $t$. Fixed-$t$ controls use the identical calibration
windows and differ only in the corruption level.

\paragraph{Generation-state reconstruction.}
We measure $E_{\mathrm{gen}}$ on 256 held-out GSM8K prompts, using one
intermediate generation state per prompt and tiling sampler steps 0--255
across prompts. Reconstruction error is evaluated on the generation
region. Confidence intervals, when reported, are obtained from 2{,}000
paired bootstrap resamples over prompts.

\paragraph{Diagnostic metrics.}
We use uncentered activation second moments, $\widehat{\Sigma}_P=M_P^{-1}\sum_{i=1}^{M_P}h_i h_i^\top$, where $M_P$ is the number of recorded activation vectors. For Fig.~\ref{fig:overview}(a), we compare the full input second moments of the block-16 query projection under clean calibration, Traj-MC calibration, and held-out generation states.
Define
$\Delta_P=(\widehat{\Sigma}_P-\widehat{\Sigma}_{\rm gen})
/\|\widehat{\Sigma}_{\rm gen}\|_F$.
We obtain an orthonormal matrix basis $(E_1,E_2)$ by applying
Gram--Schmidt to $(\Delta_{\rm clean},\Delta_{\rm traj})$
under the Frobenius inner product, and plot
$(\langle\Delta_P,E_1\rangle_F,
  \langle\Delta_P,E_2\rangle_F)$.
This places generation at the origin and exactly preserves all
three pairwise normalized Frobenius distances.
The numerical annotations report
$\|\widehat{\Sigma}_P-\widehat{\Sigma}_{\rm gen}\|_F
/\|\widehat{\Sigma}_{\rm gen}\|_F$,
with smaller values indicating closer second-moment agreement;
the arcs indicate equal mismatch.

For Fig.~\ref{fig:mc_analysis}(b), the moment-estimation error is
\[
e_\Sigma(N)=\frac{1}{|\mathcal I|}
\sum_{\ell\in\mathcal I}
\frac{\|\widehat{\Sigma}_{N,\ell}
             -\widehat{\Sigma}_{{\rm ref},\ell}\|_F}
     {\|\widehat{\Sigma}_{{\rm ref},\ell}\|_F},
\]
where $N$ counts calibration windows and $\mathcal I$ contains
the three probe layers of Table~6; the reference is independently
sampled using seed 31337.
Panel~(c) uses a separate FullGrid16 reference, averaging
16 midpoint masking levels and two mask realizations over
256 fixed C4 windows.
Its moment error $e_\Sigma^{\rm FG}(N)$ uses the same definition,
averaged over query projections at blocks 16 and 31.
For each module, the subspace distance is
\[
d_{\rm sub}(\widehat W_N,W_{\rm ref})
=\frac{\|\Pi_{\widehat W_N}-\Pi_{W_{\rm ref}}\|_F}{\sqrt{2k}},
\]
where $\Pi_W$ is the orthogonal projector onto the retained
row space of the low-rank approximation $W$, with $k=1638$.
Panel~(c) averages these distances over the same two modules;
markers and error bars summarize the mean and one standard
deviation across five mask seeds.

\section{Limitations and Discussion}
\label{sec:limitation}

\paragraph{Limitations.} Traj-MC uses a simple uniform distribution over corruption levels and independent token masking, prioritizing broad trajectory coverage without introducing sampler-specific tuning. More adaptive calibration distributions may further improve sample efficiency or alignment for particular dLLMs~\cite{yang2026target}. Our experiments focus on activation-aware low-rank compression instantiated with SVD-LLM; extending the same trajectory-aware principle to additional low-rank solvers and model families is a natural direction for future work. Finally, our analysis isolates the layerwise low-rank reconstruction problem, while its downstream consequences are evaluated empirically through generation-state fidelity and reasoning benchmarks.


\paragraph{Discussion and broader impact.}
Beyond dLLM compression, trajectory-aware calibration highlights a general design paradigm: post-training optimization for iterative generative models should align with intermediate-state dynamics rather than static, clean inputs~\cite{wen2026not}. This principle complements recent efforts to improve diffusion-model efficiency by exploiting generation dynamics, such as SADA and ZEUS~\cite{jiang2025sada,wang2026zeus}, as well as advances in low-rank model deployment, including standardized evaluation of SVD-based compression~\cite{Jiang2026JointBT} and efficient low-rank inference systems~\cite{shao2026flashsvd,wu2026flashsvd}. Together, these directions suggest that efficient deployment should jointly consider the generation process, the compression objective, and the execution substrate. By preserving more reasoning capability under a fixed compression budget, trajectory-aware calibration may provide a complementary component for efficient diffusion-based language model deployment.

\end{document}